\documentclass[runningheads]{IEEEtran}
\usepackage{bm}
\usepackage{cite}
\usepackage{amsmath}
\usepackage{array}
\usepackage{booktabs} % For pretty tables
\usepackage{caption} % For caption spacing
\usepackage{subcaption} % For sub-figures
\usepackage{graphicx}
\usepackage{pgfplots}
\usepackage[all]{nowidow}
\usepackage[utf8]{inputenc}
\usepackage{tikz}
\usepackage{multicol}
\usepackage{algpseudocode,algorithm,algorithmicx}
\newcommand{\sups}[1]{\ensuremath{^{\textrm{#1}}}}

\pgfplotsset{compat=1.14}
\begin{document}

\title{A New Sentence Extraction Strategy for Unsupervised Extractive Summarization Methods}

\author{Dehao Tao\sups{$\ast$}, Yingzhu Xiong\sups{$\ast$}, Zhongliang Yang and Yongfeng Huang
\thanks{This work is supported by the National Natural Science Foundation of China under Grant 61862002, Grant U1705261, and Grant U1936208.(Corresponding author: Yongfeng Huang.)}
\thanks{The author Dehao Tao is with the Institute for Network Sciences and Cyberspace, Tsinghua University, Beijing 100084, China (e-mail:taodh19@mails.tsinghua.edu.cn)}
\thanks{The author Yingzhu Xiong is with the Department of Computer Science, University of Wisconsin-Madison, USA (e-mail: xiong73@wisc.edu)}
\thanks{The authors Zhongliang Yang and Yongfeng Huang are with the Department of Electronic Engineering, Tsinghua University, Beijing 100084, China (e-mail:yangzl15@tsinghua.org.cn, yfhuang@mail.tsinghua.edu.cn)}
\thanks{{$\ast$} The two authors contributed equally}}

%\markboth{Journal of \LaTeX\ Class Files, Vol. 14, No. 8, August 2015}
\maketitle

\begin{abstract}
In recent years, text summarization methods have attracted much attention again thanks to the researches on neural network models. Most of the current text summarization methods based on neural network models are supervised methods which need large-scale datasets. However, large-scale datasets are difficult to obtain in practical applications. In this paper, we model the task of extractive text summarization methods from the perspective of Information Theory, and then describe the unsupervised extractive methods with a uniform framework. To improve the feature distribution and to decrease the mutual information of summarization sentences, we propose a new sentence extraction strategy which can be applied to existing unsupervised extractive methods. Experiments are carried out on different datasets, and results show that our strategy is indeed effective and in line with expectations.
% In this paper, in order to reduce the duplication within summarization, we explored unsupervised extractive summarization on reducing mutual information among summarization sentences
\end{abstract}


\IEEEpeerreviewmaketitle

\section{Introduction}

\IEEEPARstart{M}{assive} text data are generated every day with the development of the Internet. People can hardly find the proper information efficiently among such a large amount of data. Automatic text summarization, aiming to summarize the most central contents of given text with least words, has attracted more and more attention. 

There are two existing approaches to solve this problem - extractive and abstractive. In extractive summarization, sentences are chosen from the original text and then concatenated to form the summary, while in abstractive summarization, though words and phrases from the original texts may be reused, the summary is basically rewrited and tries to retain the main contents \cite{nenkova2011automatic}. Ideally, the abstractive methods can generate summarizations which are refined and in line with human habit. However, they also bring great challenge when we tried to keep the grammar of the summarizations corrrect. Abstractive summarization methods are greatly limited by text generation technology under strong semantic constraints. At present, this kind of technology can not effectively balance the naturalness, grammatical correctness and semantic accuracy of the generated text. Therefore, it is always difficult to achieve satisfactory results, which seriously restricts the practical application. In contrast, the summarization generated by extractive methods come from the original text, so the fluency and grammatical correctness of the summarization can be basically guaranteed, and the main challenge is the semantic accuracy of extracted sentences. In order to solve this challenge, previous researchers have done a 
large amount of researches.

Traditional machine learning techniques have been widely used for text summarization. Some of them used statistical features to extract summarization, such as TF-IDF \cite{shi2009study}. There are also methods using trainable systems in which models try to capture text features, such as sentence position and length \cite{radev2004centroid}, word frequency \cite{nenkova2006compositional}, and action nouns \cite{filatova2004event}. A shortcoming of these methods is that the features of summarization are always concentrating on the features the methods capture, and it may lead to the duplication of summarization. 

Neural network models are applied to text summarization in recent years \cite{nallapati2016abstractive, narayan2018don,narayan2018ranking,see2017get,zhong2020extractive,lewis2019bart}. Cheng and Lapata \cite{cheng2016neural} proposed a data-driven approach based on a neural network composed of a hierarchical document encoder and an attention-based extractor. Liu and Lapta \cite{liu2019text} treated the summary task as a binary classification problem and fine-tuned their own model based on BERT \cite{devlin2018bert}. Zhong \emph{et al} \cite{zhong2020extractive} used semantic matching to find the best summary in a set of candidate summaries. The success of these methods is inseparable from large-scale datasets which are actually hard to access in many practical applications \cite{grusky2018newsroom}. Based on this situation, researchers hope to develop unsupervised text summarization methods relying less on training data, which can play a greater role in real-world applications.

Unsupervised methods always attract researchers’ attention \cite{radev2004centroid,mihalcea2004textrank,erkan2004lexrank,xu2020unsupervised,wang2018learning}. TextRank \cite{mihalcea2004textrank} is a very popular graph-based method which scores the sentences with the graph-based ranking algorithm derived from PageRank \cite{page1999pagerank}. Mallick and Das modified TextRank by considering term frequency and inverse sentence frequency \cite{mallick2019graph}. PACSUM \cite{zheng2019sentence} changed the undirected graph into directed graph. Among these unsupervised methods, sentences are always scored according to certain rules, and the sentences with highest scores are extracted as summarization. The main problem of these methods is that, they always score the sentences with same rules, leading to the concentrated feature distribution and duplication of summarization.

To solve this problem, we describe these methods with a uniform framework, and then propose a new strategy for the sentence extraction stage of the uniform framework, which aims at avoiding repetition of summarization sentences. We made experiments on different datasets, and results show that the strategy is efficient.

\section{The proposed method}

\subsection{Task Modeling}

The most intuitive rules when we do manual text summarization is: firstly, the summarization should contain the main (but not all) contents of the article; secondly, on the basis of the first, the number of words in the summarization should be as few as possible. The two rules restrict each other. For extractive summarization methods, we take sentences as units, thus we can restate the rules as \emph{include more main contents with less sentences}. 

Let $T$ represents a text consisting of $n$ sentences \{$SE_{1}$,$SE_{2}$,\dots,$SE_{n}$\}, $SUM$ represents the summarization of $T$ consisting of $k$ sentences \{$SUM_{1}$,$SUM_{2}$,\dots,$SUM_{k}$\}. For extractive methods, $SUM$ is a subset of $T$, and the task is to maximize the information entropy of $SUM$ while minimizing sentence number $k$ of $SUM$.
Let $X$ be the random part of $SUM$, and $Y$ be the rest. According to Information Theory, the information entropy of the $SUM$ $H(SUM)$ can be calculated by
\begin{equation}
    H(SUM)=H(X,Y)= H(X) + H(Y) - I(X,Y),
    \label{eq:I}
\end{equation}
where $H(X,Y)$ is the joint entropy of $X$ and $Y$, $H(X)$ is the information entropy of $X$, $H(Y)$ is the information entropy of $Y$, and $I(X,Y)$ is the mutual information of $X$ and $Y$. We think that increasing $H(SUM)$ means \emph{include more main contents}.

Let $k$ be the number of summarization sentences. Considering the fact that there are no existing unsupervised methods which can automatically decide different $k$ for different texts, we have to regard $k$ as a set number by users rather than a variable for now.

To increase $H(SUM)$, existing methods score sentences based on textual or statistical features, and then choose the top-$k$ sentences as summarization. For instance, TextRank \cite{mihalcea2004textrank} scores sentences based on word co-occurrence, and PACSUM \cite{zheng2019sentence} scores sentences based on the order the sentences appear. 

These methods can increase $H(SUM)$ because the features they choose can screen out sentences with relatively high information entropy. However, the main problem of these unsupervised methods is that, the features (such as word frequency, position, etc.) are relatively simple and not comprehensive, and it leads to the concentrated feature distribution of summarization which also means the high $I(X,Y)$. 

We propose a new sentence extraction strategy for existing methods to improve the concentrated feature distribution and to decrease $I(X,Y)$ of summarization sentences. 
\begin{figure*}
    \centering
    \includegraphics[width=17cm]{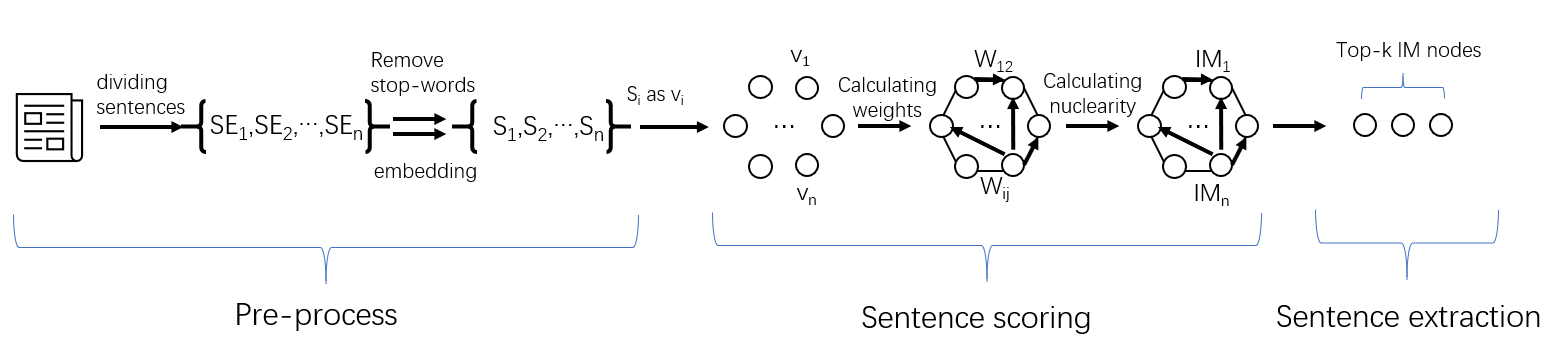}
    \caption{Framework of graph based unsupervised extractive methods}
    \label{fig:framework}
\end{figure*}
\begin{figure}
    \centering
    \includegraphics[width=5cm]{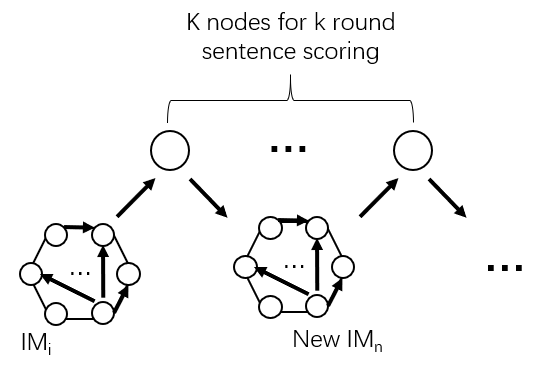}
    \caption{A new strategy to extract sentences}
    \label{fig:nu}
\end{figure}

\subsection{Decrease Mutual Information at Sentence Extraction Stage} \label{sec:parameter-estimation}

We have expounded the purpose and significance of the strategy proposed in this paper. Next, we will detail how our proposed strategy is applied in practical work. First of all, it should be clear that our work is not mutually exclusive with most existing unsupervised methods. On the contrary, our strategy is a good complement for them. To illustrate this, we first describe the methods which are suitable for our proposed strategy with a unified framework, as shown in the Fig. \ref{fig:framework}.

The first stage is pre-processing, and it converts the original text into symbol type or digital type representations. Let the representations be  \{$S_{1}$,$S_{2}$,\dots,$S_{n}$\} where $S_i$ is the representation of the $i$-th sentence. 

The second stage is sentence scoring. We use importance $IM_i$ to represent the score of the $i$-th sentence, use graph $G$ to represent a complete text, in which nodes are corresponding to sentences and edges are weighted by the similarity of every two sentences. We define such a graph with $n$ vertices as:
\begin{equation}
    \operatorname{G}=\{V,E\},
    \label{eq:G}
\end{equation}
where $V$ is a set of nodes \{$v_{1}$,$v_{2}$,\dots,$v_{n}$\}, $E$ is a set of edges $\left.\left\{e_{i, j}\right|{i,j \in[1, n]}\right\}$.
$G$ is corresponding to a complete text, $v_i$ is corresponding to $S_i$, and the weight $W_{i,j}$ for $e_{i,j}$ is calculated with $S_i$ and $S_j$. Here, the edges can be directed or undirected. A common way to calculate weights is to calculate the inner product of every two sentences by
\begin{equation}
    \operatorname{W}_{i,j}= S_{i} \cdot S_{j},
    \label{eq:n1}
\end{equation}
where $S_{i}$ and $S_{j}$ are digital type representations.
At last, importance of each node is calculated with the weights. For methods with undirected edges, importance $IM$ can be calculated by
\begin{equation}
    \operatorname{IM}_i=\lambda \cdot \sum_{j \in\{1, . ., i-1, i+1, . ., n\}} W_{i,j},
    \label{eq:n1}
\end{equation}
where $\lambda$ is a hyper-parameter. For methods with directed edges, importance $IM$ can be calculated by
\begin{equation}
     \operatorname{IM}_i=\lambda_{1} \cdot \sum_{j \in In(node_i)} W_{i,j} + \lambda_{2} \cdot \sum_{j \in Out(node_i)} W_{i,j},
     \label{eq:n2}
\end{equation}
where $\lambda_{1}$ and $\lambda_{2}$ are two hyper-parameters, $In(node_i)$ represents incoming edges of node $i$, $Out(node_i)$ represents outgoing edges of node $i$.

The last stage is sentence extraction. Most existing methods just rank the sentences according to $IM$, and select the top-$k$ sentences as summarization. As we analyzed above, the top-$k$ sentences means the concentrated feature distribution and the high $I(X,Y)$. We propose a new sentence extraction strategy to improve the concentrated feature distribution and to decrease $I(X,Y)$ at this stage. The overall process of our proposed strategy is shown in Fig. \ref{fig:nu}, and the following details how to apply it in practice.

We think high $IM$ sentences tend to have high information entropy, thus our strategy still tends to extract the high $IM$ sentences. But we do not get all the high $IM$ sentences at once like the existing methods, instead, we choose the top-1 sentence and then recompute the $IM$. 

Let the chosen sentence be the $i$-th sentence. 
To decrease the mutual information, sentences that are similar to the $i$-th sentence should be less likely extracted. Meanwhile, sentences to be extracted should still be with high information entropy. To achieve these, for the other $j$-th sentence, we decrease contribution to the calculation of $IM_{j}$ from the similarity between sentence $i$ and $j$. As mentioned above, weight $W_{i,j}$ is actually the similarity between sentence $i$ and $j$. For undirected edge, new weight $W'_{i,j}$ is recomputed by 
\begin{equation}
    W'_{i,j}= \alpha\cdot W_{i,j}, 
    \label{eq:wt}
\end{equation}
where $\alpha$ is a hyper-parameter, $i$ represents the last extracted sentence, and $j$ represents the sentence has not been extracted. For directed edge, new weight $W'_{i,j}$ is recomputed by
\begin{equation}
    W'_{i,j}= \begin{cases}
        \alpha_{1} \cdot W_{i,j} & \text{if  } i \in In(node_j) \\
        \alpha_{2} \cdot W_{i,j} & \text{otherwise}
    \end{cases},
    \label{eq:wp}
\end{equation}
where $\alpha_{1}$ and $\alpha_{2}$ are two different hyper-parameters, $i$ represents the last extracted sentence, and $j$ represents the sentence has not been extracted. 

Then we recompute $IM_j$. For methods with undirected edges, importance $IM_j$ is calculated by
\begin{equation}
    \operatorname{IM}_j=\lambda \cdot \sum_{x \notin EXT} W_{x,j}+\sum_{x \in EXT}W'_{x,j},
    \label{eq:nn1}
\end{equation}
where $\lambda$ is the same hyper-parameter from formula \ref{eq:n1}, and $EXT$ represents the set of sentences extracted. For methods with directed edges, importance $IM_j$ can be calculated by
\begin{equation}
\begin{aligned}
     \operatorname{IM}_j=\lambda_{1} \cdot  \sum_{x \in \{In(node_j)- EXT\}} \!\!\!\!\!\!\!\! W_{x,j} +\\
     \lambda_{2} \cdot  \sum_{x \in \{Out(node_j)- EXT\}} \!\!\!\!\!\!\!\!W_{x,j} + \sum_{x \in EXT}W'_{x,j},
     \label{eq:nn2}
\end{aligned}
\end{equation}
where $\lambda_1$ and $\lambda_2$ are the same hyper-parameters from formula \ref{eq:n2}, and $EXT$ represents the set of sentences extracted.

After all the $IM$ are recomputed, we rank $IM$ again and extract the next summarization sentence. The above steps are repeated until $k$ sentences are extracted. As mentioned above, the $k$ is set by users.

There are three advantages to recompute weights and importance like this. First, the more similar sentences to extracted ones are, the greater the impact will be, which is conducive to the reduction of mutual information; Second, a sentence with high information entropy but similar to extracted ones may still be extracted because it can still normally obtain importance scores from weights not affected. Third, the influence from the extracted sentences will continue in each subsequent round of sentence extraction.

For an article with n sentences, the practice of our strategy actually only costs about n-square addition operations, so the time consumption it brings is completely affordable.

\section{Experiments}

\subsection{Datasets}

In order to evaluate the strategy we proposed, we applied it to TextRank \cite{mihalcea2004textrank} and PACSUM \cite{zheng2019sentence}, and  made experiments on four different datasets, CNN/DailyMail \cite{hermann2015teaching} , NYT \cite{sandhaus2008new} , TTNews \cite{hua2017overview} and CLTS \cite{liu2020clts} . CNN/DailyMail is a commonly used news summarization dataset containing short text. New York Times (NYT) contains longer text than CNN/DailyMail. TTNews is a Chinese dataset with content from Taiwanese news media. CLTS is a newly released dataset, which contains relatively long Chinese texts.

\subsection{Deployment details}

We choose BERT \cite{devlin2018bert} as the encoder to the methods which need digital type representations. In order to see the effect of our strategy more intuitively and eliminate the influence of finetuning, we employeded the same BERT model downloaded from Internet on all the methods.

We tuned hyper-parameters on the validation set of the three datasets except the TTNews which contains only train set and test set. The best hyper-parameters are different for different methods. For instance, as shown in Fig. \ref{fig:hyper}, TextRank(our strategy) only needs one hyper-parameter $\alpha$ since the edge of it is undirected, and we chosen the best $\alpha$ = 0 for CNN/DailyMail and $\alpha$ = -0.8 for NYT. 

When evaluating on the Chinese datasets, since the ROUGE package we used does not support Chinese texts, we convert the Chinese words into Unicode format, and then use these transformed sentences as input to call ROUGE for experimental evaluation.

We set the number of extracted sentences to 3 which is a common choice for most methods.

\subsection{Result analysis}
\begin{table}[]
    \centering
    \setlength{\arrayrulewidth}{0.3mm}
    \setlength{\tabcolsep}{3.5pt}
    \renewcommand{\arraystretch}{1.1}
    \begin{tabular}{  m{2.7cm} | m{0.7cm} m{0.7cm} m{0.7cm} | m{0.7cm} m{0.7cm} m{0.7cm} }
        \hline
         Method & \multicolumn{3}{c|}{NYT} & \multicolumn{3}{c}{CNN/DailyMail} \\
         \hline
         {} & {R-1} & {R-2} & {R-L} & {R-1} & {R-2} & {R-L} \\
         \hline
         Lead3    & 35.3 & 17.2 & 31.8 & 40.2 & 17.6 & 36.5 \\
         \hline
         TextRank\cite{mihalcea2004textrank} & 32.9 & 13.2 & 28.9 & 33.1 & 12.2 & 29.7 \\
         \hline
         TextRank(our strategy) & 33.6 & 13.3 & 29.3 & 34.0 & 12.3 & 30.4 \\
         \hline
         PACSUM\cite{zheng2019sentence}   & 40.5 & 20.9 & 36.6 & 40.3 & 17.7 & 36.6 \\
         \hline
         PACSUM(our strategy)     & $\mathbf{41.3}$ & $\mathbf{21.7}$ & $\mathbf{37.4}$ & $\mathbf{40.6}$ & $\mathbf{17.7}$ & $\mathbf{36.9}$ \\
         \hline
    \end{tabular}
    \caption{ROUGE Results on NYT and CNN/DailyMail}
    \label{tab:result1}
\end{table}
\begin{table}[]
    \centering
    \setlength{\arrayrulewidth}{0.3mm}
    \setlength{\tabcolsep}{3.5pt}
    \renewcommand{\arraystretch}{1.1}
    \begin{tabular}{  m{2.7cm} | m{0.7cm} m{0.7cm} m{0.7cm} | m{0.7cm} m{0.7cm} m{0.7cm} }
    \hline
         Method & \multicolumn{3}{c|}{CLTS} & \multicolumn{3}{c}{TTNews} \\
         \hline
         {} & {R-1} & {R-2} & {R-L} & {R-1} & {R-2} & {R-L} \\
         \hline
         PACSUM\cite{zheng2019sentence}   & 41.7 & 23.4 & 41.5 & 38.8 & 20.2 & 38.5 \\
         \hline
         PACSUM(our strategy)    & $\mathbf{42.2}$ & $\mathbf{23.4}$ & $\mathbf{41.9}$ & $\mathbf{39.2}$ & $\mathbf{20.3}$ & $\mathbf{38.9}$ \\
         \hline
    \end{tabular}
    \caption{ROUGE Results on CLTS and TTNews}
    \label{tab:result2}
\end{table}
\begin{figure}
    \centering
    \includegraphics[width=4.3cm]{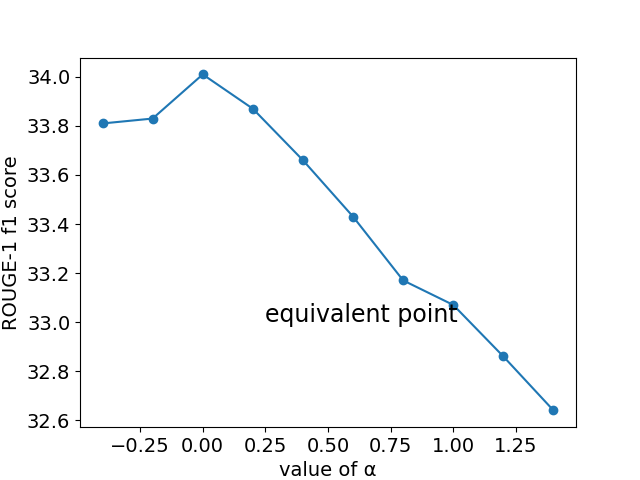}
    \includegraphics[width=4.3cm]{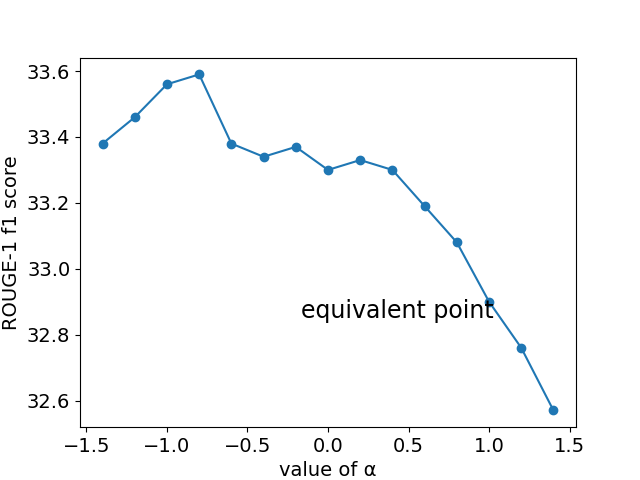}
    \caption{ROUGE-1 f1 scores of TextRank(our strategy) with different $\alpha$. $\alpha$ = $\lambda$ = 1 at equivalent point and it means TextRank is equivalent to TextRank(our strategy) there. The left figure is of CNN/DailyMail, and the right figure is of NYT.}
    \label{fig:hyper}
\end{figure}
\begin{figure}
    \centering
    \includegraphics[width=4cm]{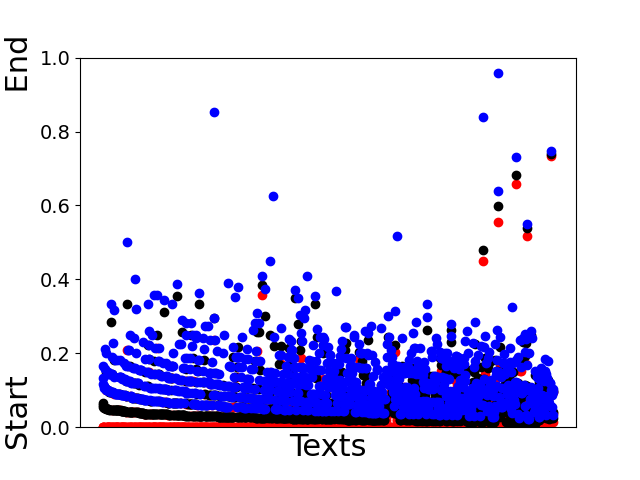}
    \includegraphics[width=4cm]{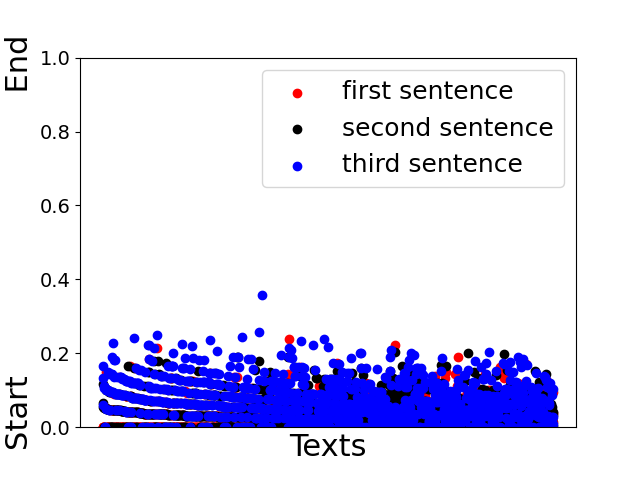}
    \includegraphics[width=4cm]{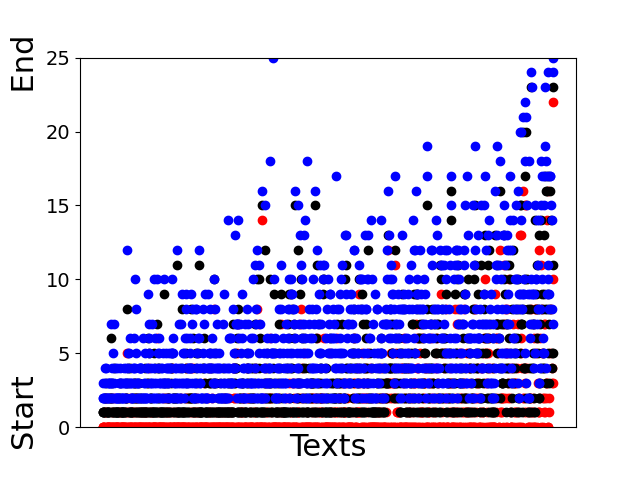}
    \includegraphics[width=4cm]{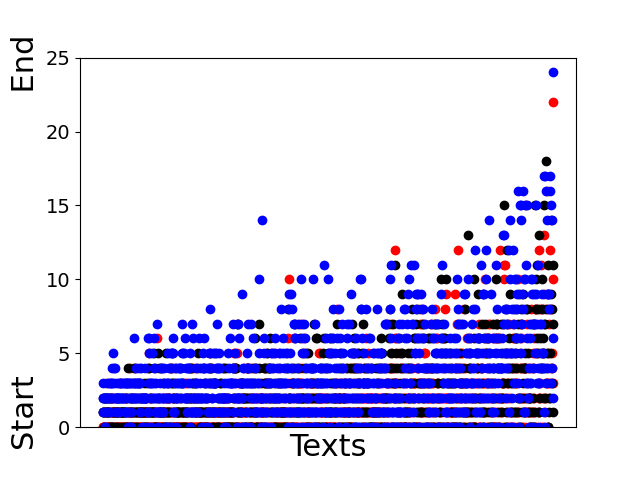}
    \caption{The position of the extracted sentences in the original text of NYT. The x-axis is the number of sentences in the original text, and the y-axis represents the position of the three sentences. The right two figures are the result of PACSUM, and the left two figures are the result of PACSUM with our strategy. The two figures above show the relative position, and the two figures below show the absolute position.}
    \label{fig:01}
\end{figure}
TextRank is a classic summarization method, and it has a great influence on PACSUM and our work. PACSUM is currently the best unsupervised text summarization method.
TextRank uses symbol type representations, and it uses TF-IDF to calculate weights, uses formula \ref{eq:n1} to calculate importance of sentences. PACSUM uses digital type representations, and it uses inner product to calculate weights, uses formula \ref{eq:n2} to calculate importance.
Also, we tested a simple method Lead3 as baseline to show the effect of other methods and the effect of our proposed strategy. Lead3 directly selects the first three sentences of the article as the summarization. Although this method is simple, it works very well in some cases, especially on the CNN/DailyMail data set.

As can be seen from the Table \ref{tab:result1}, our strategy can indeed improve the existing methods more or less. 
Lead3 method is indeed very effective on the CNN/DailyMail, and it's better than TextRank both on NYT and CNN/DailyMail.
As Table \ref{tab:result2} shows, we also made tests on CLTS and TTNews, and result shows our strategy still works.

To see if our strategy improves concentrated feature distribution of summarization, we also make statistics on the position of summarization sentences generated by PACSUM and PACSUM with our strategy. Position is the feature PACSUM uses. The results are shown in Fig. \ref{fig:01}. It can be seen that the summarization sentences extracted by PACSUM tend to focus on the front of the article, and they are so concentrated that the blue dots which represent the third sentences almost cover the red and black dots which represent the first and second sentences. Our strategy improves the position distribution obviously. 

The relation between ROUGE-1 f1 scores and hyper-parameter $\alpha$ is shown in Fig. \ref{fig:hyper}. When $\alpha = 1$, TextRank is equivalent to TextRank(our strategy). We can see a smaller $\alpha$ is more effective for both datasets. We think it's because the smaller $\alpha$ exactly decreases the mutual information among summarization sentences. We can also see that ROUGE scores will not always increase as $\alpha$ decreases. We think that's because the damage from decreasing information entropy exceeds the benefit from decreasing mutual information. We chosen the best $\alpha$ = 0 for CNN/DailyMail and $\alpha$ = -0.8 for NYT. It seems like that the weights of edges connected to extracted sentences should not contribute to the calculation of $IM$, but damage it.

\section{Conclusion}

In this paper, we describe the unified framework of unsupervised extractive summarization methods and propose a new strategy at sentence extraction stage. Our strategy aims at improving the concentrated feature distribution and decreasing the mutual information of summarization. The results of experiments confirm the effect of our strategy. We also found that our strategy decreases the mutual information of summarization sentences, meanwhile, may also decrease the information entropy to a certain extent. It may be because our method is limited to the sentence extraction stage, and a strategy throughout all the stages shall be more effective.
The average information entropy $\bar{H}$ is defined in this paper, and we believe it can help us to automatically decide different $k$ for different texts in the future.

\bibliographystyle{splncs04}
\bibliography{biblio}
\end{document}